# Tracking Extrema in Dynamic Environment using Multi-Swarm Cellular PSO with Local Search

*Somayeh Nabizadeh[1,*], Alireza Rezvanian[2], Mohammad Reza Meybodi[3]*

*Abstract*

*Many real-world phenomena can be modelled as dynamic optimization problems. In such cases, the environment problem changes dynamically and therefore, conventional methods are not capable of dealing with such problems. In this paper, a novel multi-swarm cellular particle swarm optimization algorithm is proposed by clustering and local search. In the proposed algorithm, the search space is partitioned into cells, while the particles identify changes in the search space and form clusters to create sub-swarms. Then a local search is applied to improve the solutions in the each cell. Simulation results for static standard benchmarks and dynamic environments show superiority of the proposed method over other alternative approaches.*

**Keywords:** *Dynamic Environment, Tracking Extrema, Multi Swarm Cellular PSO, Local Search.*

AUTHORS INFO

*[1*] Somayeh Nabizadeh*
  e-mail: s_nabizadeh@qiau.ac.ir
  Qazvin branch, Islamic Azad University, Qazvin, Iran

*[2] Alireza Rezvanian*
  e-mail: a.rezvanian@aut.ac.ir
  Amirkabir University of Technology (Tehran Polytechnic), Tehran, Iran

*[3] Mohammad Reza Meybodi*
  e-mail: mmeybodi@aut.ac.ir
  Amirkabir University of Technology (Tehran Polytechnic), Tehran, Iran

*Corresponding author Somayeh Nabizadeh*
  e-mail: s_nabizadeh@qiau.ac.ir
  Tel: +98-21-64545120

## I. INTRODUCTION

In many real-world problems, which are dynamic in nature, fitness function changes over time. In such cases, due to dynamic changes in the search space, conventional evolutionary algorithms are not applicable. To be more specific, an algorithm can tackle such problems if it is capable of identifying changes in the environment and finding new optimum solutions [1]. In this regard, different methods such as maintenance diversity, increased diversity, memory-based and multi-swarm methods are proposed for solving dynamic optimization problems [2]. In this research, the main focus is on a hybrid method of maintenance diversity and multi-swarm. Evolutionary algorithms such as genetic algorithm [3], differential evolution [4], artificial immune system [5,6], and ant colony optimization make these methods applicable for solving dynamic optimization problems by appropriate mechanisms.

Particle swarm optimization (PSO) algorithm has gained a significant attraction due to its simplicity and efficiency [8]. In addition, different versions of PSO are applied in dynamic environments. In the multi-swarm algorithm proposed by [9], parents maintain diversity and identify promising regions while offspring searches local areas to find local optima. In recent years, because of their satisfying results, multi-swarm algorithms, in which the particles are clustered into search groups, have got significant attention. Partly, focus of the recent works has been concentrated on methods and types of particles grouping. Recently, a new promising method based on cellular automata is proposed by *Hashemi et al.* for partitioning the solution space into cells [10,11]. In this paper, two mechanisms are proposed to maintain the diversity in cellular PSO. In the first one, clustering is used to form sub-swarms in each cell instead of searching the whole cell in order to speed up the search, whereas the second mechanism, local search is applied in each cell to improve the quality of solutions.

The rest of this paper is organized as follows: in section 2, the cellular PSO is introduced briefly. The proposed method is discussed in section 3. Section 4, provides the simulation results for static standard benchmark and dynamic environment. Finally, section 5 concludes the paper.

## II. CELLULAR PSO

The original PSO, introduced in 1990s, is based on swarm behaviour. In PSO, each solution is considered as a particle which represents a single bird in a swarm. Initially, the particles are created and positioned randomly within the search space. Afterwards, each particle is updated iteratively according to the best observed value for personal and global fitness to reach optimal fitness [12].

In Cellular PSO, the search space is partitioned and a cellular automaton (CA) is fitted to the partitioned space to maintain diversity and provide an appropriate search on the space. Each cell in the CA searches and controls its corresponding region according to some predefined rules. Each particle is assigned to a cell based on its position in the space with search procedure being performed separately for each cell and its neighbours by using the PSO. This search method provides enough diversity as well as the ability to follow multiple optimum solutions. In addition, neighbouring cells communicate information about their best known solutions which results in a more appropriate cooperation between neighbouring cells for sharing their experiences. This in turn increases efficiency of the algorithm [11].

During each iteration of the algorithm, velocity, and position of the particles are updated according to the equations below:

$$v_i(t+1) = wv_i(t) + c_1 r_1 (pBest_i - p_i(t)) + c_2 r_2 (lBestMem_k - p_i(t)) \quad (1)$$

$$p_i(t+1) = p_i(t) + v_i(t) \quad i = 1,...,m \quad (2)$$

Where $v_i$ is the velocity of the $i^{th}$ particle and $p_i$ is its position. $r_1$ and $r_2$ are uniformly distributed random variables in (0,1), while $c_1$ and $c_2$ are the learning parameters which are usually considered as equal. $w$ represents the inertia weight which may be constant or variable. $pBest_i$ denotes the best known solution for the $i^{th}$ particle and $lBestMem_k$ is the best known solution of $k^{th}$ cell neighbour to which particle $i$ belongs.

One major drawback of cellular PSO is that the number of cells increases exponentially as dimension of the problem and/or the number of the partitions increase. Moreover, it is not possible to change the number of cells during runtime. To overcome the problem of fixed number of cells, clustering is used to dynamically create groups in each cell whenever needed. By application of the clustering technique, it would be unnecessary to increase the number of cells in order to obtain a more precise search. Therefore, exponential increase in the number of cells is prevented. Furthermore, a local search procedure is applied for solution improvement.

## III. PROPOSED ALGORITHM

In cellular PSO, a CA is used for solution space partitioning. CA is known as a mathematical model of systems with several simple components which have local interactions. Using the local rules on CA, an ordered structure may be obtained from a completely random state. In CA, two well-known neighborhood structures of Von Neumann and Moore are utilized as Figure 1.

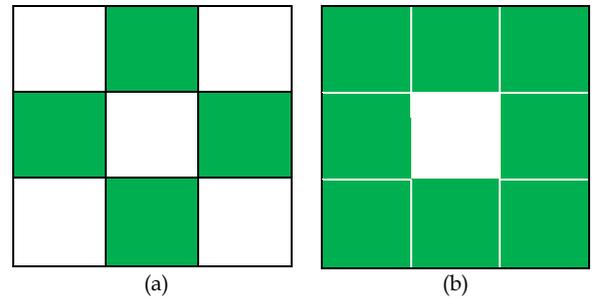

Fig. 1. 2-D Neighbourhood structure in CA; (a) Moore; (b) Von Neumann

In the proposed method, after partitioning the space into cells, clustering is generally applied to form groups of particles on which local search is applied during the cellular PSO procedure. In this algorithm each cell contains some groups, which are considered as multi-swarm having Moore neighbourhood structure.

Velocity of particles in each swam are updated as follows:

$$v_k(t+1) = a_1 r_1 (pBest_k - p_k) + a_2 r_2 (c_i^{NBest} - p_k) + wv_k(t) \quad (3)$$

Where $C_i^{Bbest}$ gives the best position in the neighbor for cell $i$. The velocity of swarm is

defined by (4):

$$p_k(t+1) = p_k(t) + v_k(t+1) \quad (4)$$

Moreover, the velocities of particles are updated in each case by equation (5).

$$v_k(t+1) = a_1 r_1 (pBest_k - p_k) + a_2 r_2 (c_i^{Best} - p_k) + w v_k(t) \quad (5)$$

In the proposed algorithm, after each change a local search is performed for each swarm which increases the efficiency of the algorithm,. The local search is applied to the $C^{Best}$ of each cell. The overall process includes definition of a magnitude and a direction of movement for each dimension to determine magnitude and direction of the search in that dimension. Moving in each dimension according to the specified magnitude and direction, fitness is calculated for the obtained position and the current position is substituted by the obtained one if improved. Otherwise, the movement direction is reversed in that dimension and a new direction is followed there. An update is implemented when the fitness is improved performing the latter action, and if not, magnitude of movement is decreased and the process begins for the next dimension. The whole procedure is performed for all dimensions until further improvement becomes impossible in all dimensions for a given movement and the minimum magnitude of movement is reached in all dimensions.

According to what discussed above, the proposed algorithm can be considered as the following steps:

1. Initialize the cells and their regions
2. Distribute the particles normally among cells in each region
3. Repeat the following steps until the termination criteria is met
   3.1. Evaluate particles
   3.2. If the change detected in the environment by memory particle
      3.2.1. Re-initialize the parameters
      3.2.2. Perform cellular movement of swarms
      3.2.3. Re-evaluate the particles
   3.3. Clustering the particles into each cell
   3.4. Update velocity and position of the particles
   3.5. Evaluate groups and cells
   3.6. Perform local search in each group
   3.7. Replace the particles in each inactive group
4. End

In the algorithm above, when particles in a group converge to a point, the group becomes inactive and its particles are used as free particles for finding better solutions in other groups of the cell or within the neighbor cells. Fig. 2 depicts the running of the algorithm and clustering of the particles in a 2-D search space.

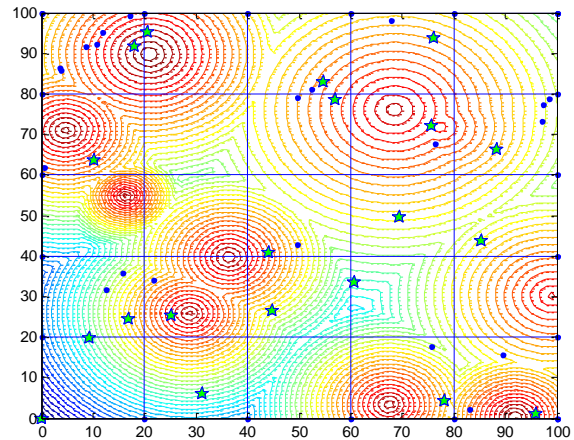

(a)

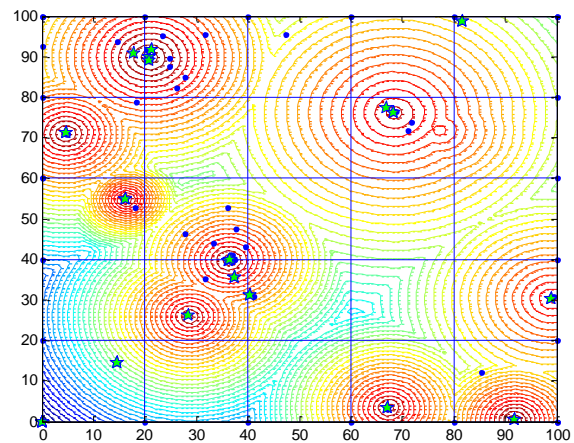

(b)

Fig. 2. (a) Initialization and search space partitioning; (b) Position of the particles in the search space after some iterations.

IV. SIMULATION RESULTS

A. Static Environments

In the first experiment, the algorithm is performed on static standard benchmark unimodal and multimodal functions including, *Sphere*, *Rastrigin*, *Griewank* and *Rosenbrock* are defined in table I [12-14].

Table I. Standard static functions for the experiments

| Name | Function | Range |
|---|---|---|
| Sphere | $f_1(x) = \sum_{i=1}^{n} x_i^2$ | $[-100,100]^D$ |
| Griewank | $f_2(x) = \sum_{i=1}^{n}(x_i^2 - 10\cos(2\pi x_i) + 10)$ | $[-5.12, 5.12]^D$ |
| Rastrigin | $f_3(x) = \frac{1}{4000}\sum_{i=1}^{n}(x_i^2) - \prod_{i=1}^{n}\cos\left(\frac{x_i}{\sqrt{i}}\right) + 1$ | $[-600,600]^D$ |
| Rosenbrock | $f_4(x) = \sum_{i=1}^{n-1}\left(100(x_{i+1} - x_i)^2\right) + 1$ | $[-5, 10]^D$ |

The experiments are accomplished assuming different dimensions of 20, 30 and 50 and population size of 3 to 5 particles in each cell by using Von Neumann neighborhood structure and 3-cell partitioning. The results for 30 independent runs of the algorithm for 1000 iterations are provided in table II, table III and table IV. The inertia weight is considered as a random variable with values between 0.4 and 0.9. A comparison of the proposed algorithm, as CPSOL, with other versions of PSO, standard PSO as SPSO [16], Fuzzy PSO as FPSO [17], Linear PSO as LPSO [18] and Robust PSO as RPSO [19] is reported.

Table II. Comparison of MCPSOL with other versions of PSO on the *Sphere* function

| Method | | Dim | | |
|---|---|---|---|---|
| | | 20 | 30 | 50 |
| SPSO | Best | 5.3606 | 14.6781 | 52.1710 |
| | Mean | 9.7219 | 20.8323 | 65.1315 |
| LPSO | Best | 2.6039 | 9.5509 | 30.2971 |
| | Mean | 4.2247 | 11.5349 | 34.0405 |
| FPSO | Best | 6.6142 | 10.0933 | 29.7984 |
| | Mean | 8.9822 | 13.5249 | 32.5106 |
| RPSO | Best | 1.4816 | 9.5509 | 12.7986 |
| | Mean | 1.9959 | 11.5349 | 16.5475 |
| CPSOL | Best | 0.9351 | 2.0064 | 8.5612 |
| | Mean | 1.9617 | 3.7861 | 14.3182 |

Table III. Comparison of MCPSOL with other versions of PSO on the *Rastrigin* function

| Method | | Dim | | |
|---|---|---|---|---|
| | | 20 | 30 | 50 |
| SPSO | Best | 67.3994 | 133.3642 | 367.5225 |
| | Mean | 110.6389 | 153.4576 | 404.0451 |
| LPSO | Best | 137.4023 | 147.3715 | 351.7914 |
| | Mean | 142.6308 | 155.2974 | 369.1274 |
| FPSO | Best | 102.2786 | 146.6628 | 301.9003 |
| | Mean | 115.1138 | 157.0243 | 320.5474 |
| RPSO | Best | 64.7160 | 131.3496 | 296.5793 |
| | Mean | 73.2037 | 144.1901 | 316.9913 |
| CPSOL | Best | 19.4109 | 43.8256 | 65.3681 |
| | Mean | 31.0693 | 179.2361 | 227.3218 |

Table IV. Comparison of MCPSOL with other versions of PSO on the *Griewank* function

| Method | | Dim | | |
|---|---|---|---|---|
| | | 20 | 30 | 50 |
| SPSO | Best | 117.7599 | 318.8507 | 503.0944 |
| | Mean | 174.0772 | 339.5614 | 702.5057 |
| LPSO | Best | 159.9319 | 339.6826 | 715.4197 |
| | Mean | 216.4355 | 395.9068 | 837.5857 |
| FPSO | Best | 159.6489 | 342.8425 | 643.2599 |
| | Mean | 198.4451 | 405.9346 | 827.6388 |
| RPSO | Best | 178.3643 | 342.0737 | 664.6935 |
| | Mean | 209.1941 | 426.9451 | 780.3784 |
| CPSOL | Best | 1.3182 | 3.4048 | 19.0432 |
| | Mean | 2.6793 | 7.5687 | 24.3255 |

Table V. Comparison of MCPSOL with other versions of PSO on the *Rosenbrock* function

| Method | | Dim | | |
|---|---|---|---|---|
| | | 20 | 30 | 50 |
| SPSO | Best | 122.5061 | 24105.353 | 139662.26 |
| | Mean | 106904.17 | 107219.21 | 318569.65 |
| LPSO | Best | 1222.7753 | 6874.5738 | 206034.62 |
| | Mean | 105257.12 | 71610.811 | 355794.89 |
| FPSO | Best | 805.5753 | 19582.926 | 75437.949 |
| | Mean | 109189.95 | 100901.54 | 271848.10 |
| RPSO | Best | 629.0278 | 3530.0328 | 6531.0425 |
| | Mean | 10229.24 | 78676.722 | 158941.32 |
| CPSOL | Best | 180.0802 | 284.4018 | 4843239 |
| | Mean | 496.9682 | 845.4121 | 1042.18 |

*S. Nabizadeh, A. Rezvanian, M. R. Meybodi*

## B. Dynamic Environments

In order to evaluate the proposed algorithm in dynamic environments, several experiments performed on two famous dynamic environments as moving parabolic function and moving peaks benchmarks.

### B.1. Experiments on moving parabolic function

In the first experiment, order to evaluate the proposed method in dynamic environment, dynamic moving parabolic function generator, developed by *Angeline* [20] is employed, which is illustrated in figure 3. A moving parabolic benchmark changes by Δk using the following equation in this dynamic environment,:

$$f(x,y,z) = x^2 + y^2 + z^2 \qquad (6)$$

Where, according to the movements one may consider the equation:

$$f(x) = \sum_{i=1}^{3}(x_i - \Delta k)^2 \qquad (7)$$

The movements are linear, circular or Gaussian with a magnitude of $\tau$ and frequency of $f$ satisfying the following equations.

$$\Delta k = \Delta k + \tau \qquad (8)$$

$$\Delta k = \begin{cases} \Delta k + \tau \sin\left(\dfrac{2\pi t}{25}\right) & k \in oven \\ \Delta k + \tau \cos\left(\dfrac{2\pi t}{25}\right) & k \in odd \end{cases} \qquad (9)$$

$$\Delta k = \Delta k + N(0,1) \qquad (10)$$

Where *t* in equation (9) denotes the cumulative number of changes in the function.

Different types of changes are used in the experiments with d=30, f=200, 1000 and τ=0.01, 0.1. The dynamic moving parabolic is applied to Sphere function in the interval [-50, 50].

In order to compare the proposed method with other algorithms, the offline error (OE), provided by the equation (12) is used [15].

$$OE = \frac{1}{T}\sum_{t=1}^{T}\left(f(p_{best}(t))\right) \qquad (11)$$

Where, *f* is the fitness function, *T* represents the maximum number of iterations and *pBest(t)* is the best known global solution found by the algorithm in iteration *t*.

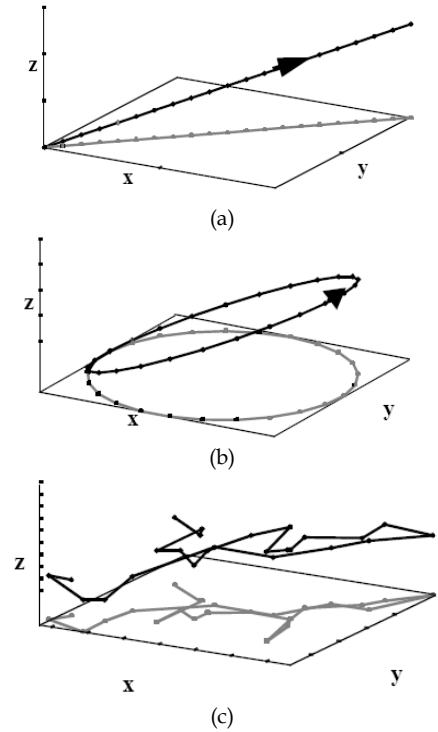

Fig. 3. Example dynamics; (a) Linear dynamic; (b) Circular dynamic; (c) Gaussian dynamic [20].

In this experiment, the proposed algorithm as CPSOL is compared with RPSO [21], mQSO 10(5+1q) [22], AmQSO [23] and CPSO [11] by offline error. For each one of the three different movements the results of OE are provided in Table VI, VII, and VIII.

Table VI. Offline error and standard deviation in dynamic environment for *Linear* movement

| F | τ | AmQSO | mQSO | RPSO | CPSO | CPSOL |
|---|---|---|---|---|---|---|
| 200 | 0.01 | 133.48±2.56 | 99.36±2.84 | 20.57±0.10 | 33.78±1.08 | 33.24±2.03 |
|  | 0.1 | 189.63±2.20 | 100.12±3.11 | 22.85±0.13 | 33.59±0.91 | 33.16±1.91 |
| 1000 | 0.01 | 27.17±0.48 | 20.04±0.63 | 0.81±0.01 | 9.18±0.19 | 8.95±0.57 |
|  | 0.1 | 90.56±0.78 | 20.11±0.64 | 0.78±0.01 | 10.70±0.22 | 9.87±0.39 |

Table VII. Offline error and standard deviation in dynamic environment for *Circular* movement

| F | τ | AmQSO | mQSO | RPSO | CPSO | CPSOL |
|---|---|---|---|---|---|---|
| 200 | 0.01 | 134.65±2.52 | 95.73±2.64 | 25.10±0.11 | 32.94±1.12 | 33.35±1.23 |
|  | 0.1 | 132.51±2.49 | 98.07±3.08 | 24.80±0.10 | 33.86±0.89 | 33.63±1.14 |
| 1000 | 0.01 | 26.92±0.49 | 19.69±0.65 | 0.82±0.01 | 7.45±0.17 | 7.37±0.34 |
|  | 0.1 | 27.89±0.45 | 19.48±0.67 | 0.81±0.01 | 9.27±0.19 | 8.94±0.32 |

Table VIII. Offline error and standard deviation in dynamic environment for *Gaussian* movement

| F | τ | AmQSO | mQSO | RPSO | CPSO | CPSOL |
|---|---|-------|------|------|------|-------|
| 200 | 0.01 | 133.65±2.40 | 98.26±3.09 | 25.10±0.11 | 33.41±1.06 | 33.06±1.19 |
|  | 0.1 | 134.60±2.60 | 99.71±3.20 | 24.90±0.13 | 33.56±1.14 | 33.29±1.08 |
| 1000 | 0.01 | 27.15±0.46 | 19.94±0.62 | 0.82±0.01 | 7.03±0.18 | 6.85±1.11 |
|  | 0.1 | 27.61±0.51 | 19.83±0.66 | 0.82±0.01 | 8.87±0.17 | 8.62±1.24 |

The proposed method is superior to original Cellular PSO for all three types of movements while RPSO has the best performance among all the existing algorithms and provides more satisfying results. Generally, the proposed algorithm demonstrates acceptable performance in comparison with the original PSO.

### B.2. Experiments on moving peaks benchmark

In the second experiment, In order to evaluate the proposed algorithm in dynamic environments, several experiments are performed on Moving Peaks Benchmark (MPB). In the MPB, there are some peaks in a multi-dimensional space, where the height, width, and position of each peak alter when the environment changes. Unless stated otherwise, the parameters of MPB are set to the values listed in table 1 [4, 11].

Table IX. Default settings of MPB

| Parameter | Value |
|---|---|
| number of peaks $m$ | 10 |
| Frequency of change $f$ | every 5000 evaluations |
| height severity | 7.0 |
| width severity | 1.0 |
| peak shape | Cone |
| shift length $s$ | 1.0 |
| number of dimensions $D$ | 5 |
| cone height range $H$ | [30.0, 70.0] |
| cone width range $W$ | [1, 12] |
| cone standard height $I$ | 50.0 |
| Search space range $A$ | [0, 100] |

For the proposed method the inertia weight is considered as a random variable between 0.4 and 0.9. The acceleration coefficient is set to 1.496180, the number of particles is 40; the type of neighborhood structure is Moore and the size of partition is 5.

In these experiments, proposed algorithm so called multi swarm cellular PSO based on local search as CPSOCL is compared with Hibernating Multi Swarm Optimization as (HmSO) [24], Learning Automata based Immune Algorithm as (LAIA) [5], Cellular Differential Evolution as (CDE) [4], Cellular Particle Swarm Optimization as (CPSO) [11], by offline error. For each experiment, the average offline error and standard deviation of 30 time-independent runs is addressed. The results of several dynamics are also listed in the table X, to XIII.

Table X. Offline Error ± Standard Error for F=500

| M | HmSO | LAIA | CDE | CPSO | CPSOL |
|---|------|------|-----|------|-------|
| 1 | 8.53±0.49 | 7.34±0.32 | 8.20±0.19 | 7.81±0.51 | 8.29±0.55 |
| 5 | 7.40±0.31 | 7.05±0.39 | 6.06±0.05 | 6.59±0.31 | 6.29±0.21 |
| 10 | 7.56±0.27 | 6.91±0.32 | 5.93±0.04 | 7.35±0.22 | 5.45±0.17 |
| 20 | 7.81±0.20 | 6.95±0.38 | 5.60±0.03 | 7.79±0.27 | 5.47±0.19 |
| 30 | 8.33±0.18 | 6.92±0.33 | 5.56±0.03 | 7.88±0.23 | 5.59±0.12 |
| 40 | 8.45±0.18 | 6.84±0.31 | 5.47±0.02 | 7.83±0.21 | 5.63±0.16 |
| 50 | 8.83±0.17 | 6.43±0.29 | 5.47±0.02 | 8.12±0.22 | 5.74±0.11 |
| 100 | 8.85±0.16 | 6.58±0.26 | 5.29±0.02 | 7.90±0.24 | 5.45±0.07 |
| 200 | 8.85±0.16 | 6.41±0.27 | 5.07±0.02 | 7.82±0.20 | 5.79±0.10 |

Table XI. Offline Error ± Standard Error for F=1000

| M | HmSO | LAIA | CDE | CPSO | CPSOL |
|---|------|------|-----|------|-------|
| 1 | 4.46±0.26 | 4.96±0.32 | 4.98±0.35 | 5.86±0.42 | 4.74±0.32 |
| 5 | 4.27±0.08 | 4.01±0.31 | 3.96±0.04 | 5.26±0.26 | 3.95±0.21 |
| 10 | 4.61±0.07 | 3.94±0.29 | 3.98±0.03 | 5.75±0.23 | 3.20±0.20 |
| 20 | 4.66±0.12 | 3.72±0.29 | 4.53±0.02 | 5.74±0.19 | 3.52±0.17 |
| 30 | 4.83±0.09 | 4.03±0.31 | 4.77±0.02 | 5.84±0.16 | 3.96±0.12 |
| 40 | 4.82±0.09 | 3.97±0.32 | 4.87±0.02 | 5.84±0.17 | 4.21±0.17 |
| 50 | 4.96±0.03 | 4.22±0.31 | 4.87±0.02 | 5.84±0.14 | 3.98±0.11. |
| 100 | 5.14±0.08 | 4.19±0.32 | 4.85±0.02 | 5.73±0.11 | 4.13±0.12 |
| 200 | 5.25±0.08 | 4.38±0.31 | 4.46±0.01 | 5.48±0.11 | 4.15±0.01 |

Table XII. Offline Error ± Standard Error for F=2500

| M | HmSO | LAIA | CDE | CPSO | CPSOL |
|---|---|---|---|---|---|
| 1 | 1.75±0.10 | 2.48±0.15 | 2.38±0.78 | 3.78±0.25 | 2.31±0.21 |
| 5 | 1.92±0.11 | 2.51±0.19 | 2.12±0.02 | 2.91±0.14 | 2.01±0.13 |
| 10 | 2.39±0.16 | 2.82±0.27 | 2.42±0.02 | 3.18±0.16 | 1.56±0.15 |
| 20 | 2.46±0.09 | 3.16±0.36 | 3.05±0.04 | 3.65±0.13 | 2.41±0.13 |
| 30 | 2.57±0.05 | 3.14±0.33 | 3.29±0.03 | 3.90±0.11 | 2.78±0.10 |
| 40 | 2.56±0.06 | 3.02±0.31 | 3.43±0.03 | 4.20±0.13 | 2.90±0.12 |
| 50 | 2.65±0.05 | 3.05±0.31 | 3.44±0.02 | 4.08±0.11 | 3.18±0.09 |
| 100 | 2.72±0.04 | 3.14±0.35 | 3.36±0.01 | 4.23±0.09 | 3.22±0.07 |
| 200 | 2.81±0.04 | 3.08±0.32 | 3.13±0.01 | 4.09±0.10 | 3.09±0.12 |

Table XIII. Offline Error ± Standard Error for F=5000

| M | HmSO | LAIA | CDE | CPSO | CPSOL |
|---|---|---|---|---|---|
| 1 | 0.87±0.05 | 1.94±0.19 | 1.53±0.07 | 2.36±0.14 | 1.02±0.14 |
| 5 | 1.18±0.04 | 2.09±0.18 | 1.50±0.04 | 1.94±0.16 | 0.99±0.15 |
| 10 | 1.42±0.04 | 2.14±0.15 | 1.64±0.03 | 2.09±0.13 | 1.75±0.10 |
| 20 | 1.50±0.06 | 2.97±0.21 | 2.64±0.05 | 2.94±0.13 | 1.93±0.11 |
| 30 | 1.65±0.04 | 2.98±0.23 | 2.62±0.05 | 3.04±0.09 | 2.28±0.10 |
| 40 | 1.65±0.05 | 3.07±0.29 | 2.76±0.05 | 3.16±0.11 | 2.62±0.09 |
| 50 | 1.66±0.02 | 2.93±0.27 | 2.75±0.05 | 3.19±0.10 | 2.74±0.10 |
| 100 | 1.68±0.03 | 3.06±0.24 | 2.73±0.03 | 3.24±0.09 | 2.84±0.12 |
| 200 | 1.71±0.02 | 2.95±0.23 | 2.61±0.02 | 3.15±0.08 | 2.69±0.08 |

According to the results of the table X to XIII, the proposed algorithm is relatively advantageous over alternative algorithms.

## V. CONCLUSIONS

In this paper, an extension of cellular PSO algorithm augmented by clustering and local search in cellular environment is proposed. The inspiration for this research was to perform a more precise search without increasing the number of partitions. This is obtained by defining and using groups in each cell. The simulation results on both static and dynamic environments reveal an improvement as compared with its original version.